\documentclass{article}
\usepackage{ijcai19}

\usepackage{times}
\usepackage{soul}
\usepackage{url}
\usepackage[hidelinks]{hyperref}
\usepackage[utf8]{inputenc}
\usepackage[small]{caption}
\usepackage{graphicx}
\usepackage{amsmath}
\usepackage{booktabs}
\usepackage[T1]{fontenc}
\usepackage[inline]{enumitem}

\usepackage{blindtext}

\usepackage{emptypage}

\usepackage{graphicx}

\usepackage{float}

\usepackage{amsmath}
\usepackage{amssymb}

\usepackage{gensymb}
\usepackage{textcomp}

\usepackage{setspace}

\usepackage{multirow}

\usepackage[format=hang,font=small,labelfont=bf]{caption}

\usepackage{etoolbox}

\usepackage{lipsum}

\usepackage{xcolor}
\colorlet{dark-blue}{blue!50!black}
\colorlet{dark-cyan}{cyan!75!black}
\colorlet{dark-purple}{purple!50!black}
\colorlet{dark-red}{red!75!black}
\colorlet{dark-green}{green!75!black}
\colorlet{dark-orange}{orange!50!black}
\colorlet{dark-gray}{black!75}
\colorlet{light-gray}{black!30}
\colorlet{hidden}{light-gray}
\colorlet{todo}{red!85!black}
\colorlet{todoref}{purple!70!black}
\definecolor{nice-red}{HTML}{E41A1C}
\definecolor{nice-orange}{HTML}{FF7F00}
\definecolor{nice-yellow}{HTML}{FFC020}
\definecolor{nice-green}{HTML}{4DAF4A}
\definecolor{nice-blue}{HTML}{377EB8}
\definecolor{nice-purple}{HTML}{984EA3}

\usepackage[xindy]{glossaries}

\usepackage{booktabs}

\usepackage{tikz}
\usetikzlibrary{calc,trees,positioning,arrows,chains,shapes.geometric,%
  decorations.pathreplacing,decorations.pathmorphing,shapes,%
  matrix,shapes.symbols,fit,decorations,arrows.meta}

\PassOptionsToPackage{hyphens}{url}\usepackage{hyperref}
\hypersetup{
 unicode=false,          %
 pdftoolbar=true,        %
 pdfmenubar=true,        %
 pdffitwindow=false,     %
 pdfstartview={FitH},    %
 pdfauthor={Tim Rocktäschel},     %
 pdfnewwindow=true,      %
 colorlinks=true,        %
 linkcolor=black,        %
 citecolor=blue!50!black,    %
 filecolor=black,        %
 urlcolor=black          %
}

\usepackage{subcaption}

\usepackage{microtype}

\usepackage{url}

\usepackage{soul}

\usepackage{changepage}

\usepackage{xargs}                      %
 
\usepackage{bm}

\usepackage[capitalize]{cleveref} %
\crefformat{equation}{Eq.~#2#1#3}
\Crefformat{equation}{Equation~#2#1#3}
\crefrangeformat{equation}{Eqs.~#3#1#4 to~#5#2#6}
\Crefrangeformat{equation}{Equations~#3#1#4 to~#5#2#6}
\crefmultiformat{equation}{Eqs.~#2#1#3}%
{ and~#2#1#3}{, #2#1#3}{ and~#2#1#3}
\Crefmultiformat{equation}{Equations~#2#1#3}%
{ and~#2#1#3}{, #2#1#3}{ and~#2#1#3}

\usepackage{stmaryrd} %

\usepackage[outline]{contour}

\usepackage[colorinlistoftodos,prependcaption,textsize=small]{todonotes}

\usepackage{xcolor}
\colorlet{dark-blue}{blue!50!black}
\colorlet{dark-cyan}{cyan!75!black}
\colorlet{dark-purple}{purple!50!black}
\colorlet{dark-red}{red!75!black}
\colorlet{dark-green}{green!75!black}
\colorlet{dark-orange}{orange!50!black}
\colorlet{dark-gray}{black!75}
\colorlet{light-gray}{black!30}
\colorlet{hidden}{light-gray}
\colorlet{todo}{red!85!black}
\colorlet{todoref}{purple!70!black}
\definecolor{nice-red}{HTML}{E41A1C}
\definecolor{nice-orange}{HTML}{FF7F00}
\definecolor{nice-yellow}{HTML}{FFC020}
\definecolor{nice-green}{HTML}{4DAF4A}
\definecolor{nice-blue}{HTML}{377EB8}
\definecolor{nice-purple}{HTML}{984EA3}

\usepackage{xargs}

\newcounter{trCounter}
\newif\iftrvar
\trvartrue
\iftrvar
\newcommand{\tr}[1]{{\small \color{blue} \refstepcounter{trCounter}\textsf{[TR]$_{\arabic{trCounter}}$:{#1}}}}
\else
\newcommand{\tr}[1]{}
\fi

\newcounter{nnCounter}
\newif\ifnnvar
\nnvartrue
\ifnnvar
\newcommand{\nn}[1]{{\small \color{orange} \refstepcounter{nnCounter}\textsf{[NN]$_{\arabic{nnCounter}}$:{#1}}}}
\else
\newcommand{\nn}[1]{}
\fi

\newcounter{jlCounter}
\newif\ifjlvar
\jlvartrue
\ifjlvar
\newcommand{\jl}[1]{{\small \color{cyan} \refstepcounter{jlCounter}\textsf{[JL]$_{\arabic{jlCounter}}$:{#1}}}}
\else
\newcommand{\jl}[1]{}
\fi

\newcounter{swCounter}
\newif\ifswvar
\swvartrue
\ifswvar
\newcommand{\sw}[1]{{\small \color{red} \refstepcounter{swCounter}\textsf{[SW]$_{\arabic{swCounter}}$:{#1}}}}
\else
\newcommand{\sw}[1]{}
\fi

\newcounter{gfCounter}
\newif\ifgfvar
\gfvartrue
\ifgfvar
\newcommand{\gf}[1]{{\small \color{olive} \refstepcounter{gfCounter}\textsf{[GF]$_{\arabic{gfCounter}}$:{#1}}}}
\else
\newcommand{\gf}[1]{}
\fi

\newcounter{jfCounter}
\newif\ifjfvar
\jfvartrue
\ifjfvar
\newcommand{\jf}[1]{{\small \color{purple} \refstepcounter{jfCounter}\textsf{[JF]$_{\arabic{jfCounter}}$:{#1}}}}
\else
\newcommand{\jf}[1]{}
\fi

\newcounter{etgCounter}
\newif\ifetgvar
\etgvartrue
\ifetgvar
\newcommand{\etg}[1]{{\small \color{green!50!black} \refstepcounter{etgCounter}\textsf{[ETG]$_{\arabic{etgCounter}}$:{#1}}}}
\else
\newcommand{\etg}[1]{}
\fi

\iftrue
\renewcommand{\jl}[1]{}
\renewcommand{\nn}[1]{}
\renewcommand{\tr}[1]{}
\renewcommand{\gf}[1]{}
\renewcommand{\jf}[1]{}
\renewcommand{\etg}[1]{}
\renewcommand{\sw}[1]{}
\fi

\usepackage{xcolor}
\colorlet{dark-blue}{blue!50!black}
\colorlet{dark-cyan}{cyan!75!black}
\colorlet{dark-purple}{purple!50!black}
\colorlet{dark-red}{red!75!black}
\colorlet{dark-green}{green!75!black}
\colorlet{dark-orange}{orange!50!black}
\colorlet{dark-gray}{black!75}
\colorlet{light-gray}{black!30}
\colorlet{hidden}{light-gray}
\colorlet{todo}{red!85!black}
\colorlet{todoref}{purple!70!black}
\definecolor{nice-red}{HTML}{E41A1C}
\definecolor{nice-orange}{HTML}{FF7F00}
\definecolor{nice-yellow}{HTML}{FFC020}
\definecolor{nice-green}{HTML}{4DAF4A}
\definecolor{nice-blue}{HTML}{377EB8}
\definecolor{nice-purple}{HTML}{984EA3}

\newcommand{\eg}{\emph{e.g.}}
\newcommand{\ie}{\emph{i.e.}}

\makeatletter
\newcommand*\bdot{\mathpalette\bdot@{.5}}
\newcommand*\bdot@[2]{\mathbin{\vcenter{\hbox{\scalebox{#2}{$\m@th#1\bullet$}}}}}
\makeatother

\newcommand{\module}[1]{\verb~#1~}

\newcommand{\be}{\begin{equation}}
\newcommand{\ee}{\end{equation}}
\newcommand{\bea}{\begin{eqnarray}}
\newcommand{\eea}{\end{eqnarray}}
\newcommand{\beaa}{\begin{eqnarray*}}
\newcommand{\eeaa}{\end{eqnarray*}}

\DeclareMathAlphabet{\mathpzc}{OT1}{pzc}{m}{n}

\newacronym{AUC}{AUC}{Area Under Curve}
\newacronym{CNN}{CNN}{Convolutional Neural Network}
\newacronym{GRU}{GRU}{Gated Recurrent Unit}
\newacronym{ILP}{ILP}{Inductive Logic Programming}
\newacronym{KB}{KB}{Knowledge Base}
\newacronym[longplural={Long Short-Term Memorie Units}]{LSTM}{LSTM}{long short-term memory}
\newacronym{NLP}{NLP}{Natural Language Processing}
\newacronym{NLU}{NLU}{Natural Language Understanding}
\newacronym{NLI}{NLI}{Natural Language Inference}
\newacronym{NTP}{NTP}{Neural Theorem Prover}
\newacronym{NTN}{NTN}{Neural Tensor Network}
\newacronym{MAP}{MAP}{Mean Average Precision}
\newacronym{MRR}{MRR}{Mean Reciprocal Rank}
\newacronym{MLP}{MLP}{Multi-layer Perceptron}
\newacronym{OpenIE}{OpenIE}{Open Information Extraction}
\newacronym{PCA}{PCA}{Principal Component Analysis}
\newacronym{QRNN}{QRNN}{Quasi-Recurrent Neural Networks}
\newacronym{RBF}{RBF}{Radial Basis Function}
\newacronym{RL}{RL}{Reinforcement Learning}
\newacronym{RNN}{RNN}{Recurrent Neural Network}
\newacronym{RTE}{RTE}{Recognizing Textual Entailment}
\newacronym{SGD}{SGD}{Stochastic Gradient Descent}
\newacronym{SNLI}{SNLI}{Stanford Natural Language Inference}

\newcommand{\citep}[1]{\cite{#1}}
\newcommand{\citet}[1]{\cite{#1}}

\title{A Survey of Reinforcement Learning Informed by Natural Language}

\author{
Jelena Luketina$^{1}$\footnote{Contact authors: jelena.luketina@cs.ox.ac.uk, rockt@fb.com.}\and
Nantas Nardelli$^{1,2}$\and
Gregory Farquhar$^{1,2}$\and
Jakob Foerster$^{2}$\and\\
Jacob Andreas$^{3}$\and
Edward Grefenstette$^{2,4}$\and
Shimon Whiteson$^{1}$\and
Tim Rockt{\"a}schel$^{2,4*}$
\affiliations
$^1$University of Oxford\and
$^2$Facebook AI Research\\
$^3$Massachusetts Institute of Technology\and
$^4$University College London\\
}

\begin{document}

\maketitle

\begin{abstract}
To be successful in real-world tasks, \gls{RL} needs to exploit the compositional, relational, and hierarchical structure of the world, and learn to transfer it to the task at hand.
Recent advances in representation learning for language make it possible 
to build models that acquire world knowledge from text corpora and integrate this knowledge into downstream decision making problems. We thus argue that the time is right to investigate a tight integration of natural language understanding into \gls{RL} in particular. %
We survey the state of the field, including work on instruction following,
text games, %
and learning from textual domain knowledge.
Finally, we call for the development of new environments as well as further investigation into the potential uses of recent \gls{NLP} techniques for such tasks. 

\jl{}\nn{}\tr{}\gf{}\jf{}\sw{}\etg{}
\end{abstract}
\section{Introduction}

 \begin{figure*}[t!]
     \centering
     \includegraphics[width=\textwidth]{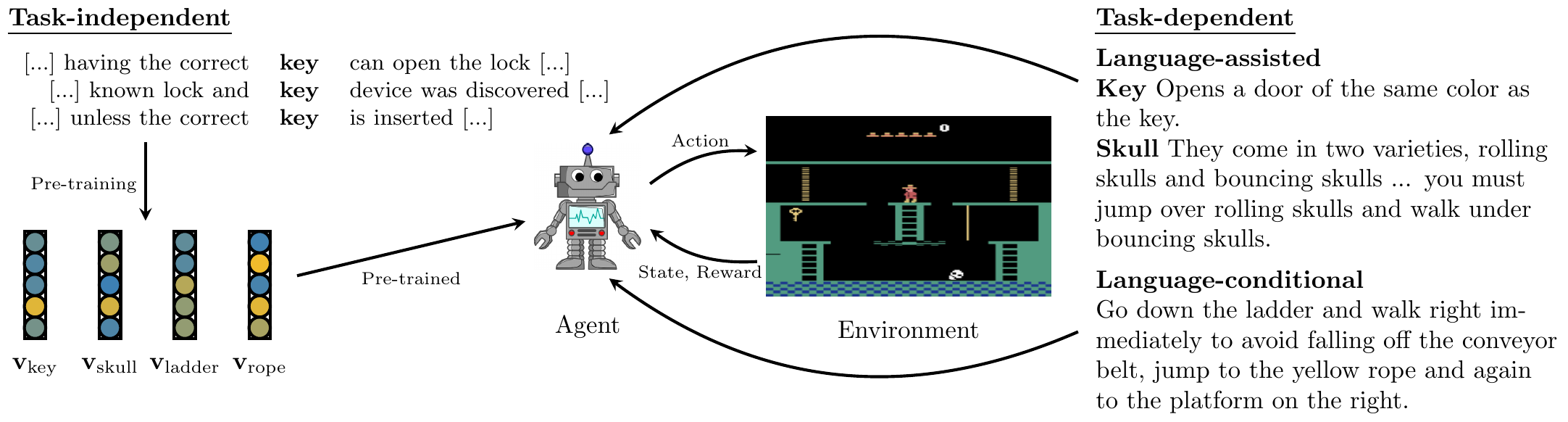}
     \caption{Illustration of different roles and types of natural language information in reinforcement learning. We differentiate between \emph{language-conditional} setting in which language is a part of the task formulation (\eg{} natural language instructions that specify the goal or reward), and \emph{language-assisted} setting where language information is not necessary to solve the task but can assists learning (\eg{} by providing information about the environment dynamics). The language information itself can be \emph{task-dependent}, \ie{} specific to the task such as tutorials or instructions, or \emph{task-independent}, for instance, conveying general priors about the world through pre-trained language representations.}
     \label{fig:illustration}
 \end{figure*}

Languages, whether natural or formal, allow us to encode abstractions, to generalize, to communicate plans, intentions, and requirements, both to other parties and to ourselves~\citep{gopnik1987development}.
These are fundamentally desirable capabilities of artificial agents.
However, agents trained with traditional approaches within dominant paradigms such as Reinforcement Learning (RL) and Imitation Learning (IL) typically lack such capabilities, and struggle to efficiently learn from interactions with rich and diverse environments.
In this paper, we argue that the time has come for natural language to become a first-class citizen of solutions to sequential decision making problems (\ie{} those often approached with \gls{RL}\footnote{We write RL for brevity and to focus on a general case, but our arguments are relevant for many sequential decision making approaches, including IL and planning.}).
We survey recent work and tools that are beginning to make this shift possible, and outline next research steps.

Humans are able to learn quickly in new environments due to a rich set of commonsense priors about the world \citep{spelke2007core}, some of which are reflected in natural language \citep{shusterman2011cognitive,DBLP:conf/aaaiss/TsividisPXTG17}.
It is thus reasonable to question whether agents can learn not only
from rewards or demonstrations, but also from information encoded using language, in order to improve generalization and sample efficiency---especially when (a) learning is severely constrained by data efficiency due to limited or expensive environment interactions, and where (b) human priors that would help to solve the task are or can be expressed easily in natural language. 
Furthermore, many, if not most, real-world tasks require agents to process language by design, whether to enable interaction with humans, or when using existing interfaces and knowledge bases. This suggests that the use of language in \gls{RL} has both broad and important applications.

Information contained in both generic and task-specific large textual corpora may be highly valuable for decision making.
Pre-training neural representations of words and sentences from large generic corpora has already shown great success in transferring syntactic and, to some extent, semantic information to downstream tasks in natural language understanding \citep{DBLP:conf/emnlp/PetersNZY18,DBLP:journals/corr/abs-1901-05287,tenney2018what}.
Cross-domain transfer from self-supervision on generic language data might similarly help to initialize RL agents.
Task-specific corpora like wikis or game manuals could be leveraged by machine reading techniques \citep{banko2007open} to inform agents of valuable features and policies \citep{eisenstein_reading_2009,branavan_learning_2012}, or task-specific environmental dynamics and reward structures \citep{narasimhan_grounding_2017,bahdanau_learning_2018}.

Previous attempts at using language for \gls{RL} tasks in these ways have mostly been limited to relatively small corpora~\citep{janner_representation_2017}, or synthetic language \citep{hermann_grounded_2017}.
We argue that recent advances in representation learning \citep{DBLP:conf/naacl/PetersNIGCLZ18,devlin_bert:_2018,radford2019language} make it worth revisiting this research agenda with a much more ambitious scope.
While the problem of grounding (\ie{} learning the correspondence between language and environment features) remains a significant research challenge, past work has already shown that high-quality linguistic representations can assist cross-modal transfer outside the context of \gls{RL} (\eg{} using semantic relationships between labels to enable zero-shot transfer in image classification \citep{frome_devise:_2013,socher_zero-shot_2013}).

We first provide background on \gls{RL} and techniques for self-supervision and transfer in \gls{NLP} (\S\ref{sec:background}).
We then review prior work, considering settings where interaction with language is necessary (\S\ref{sec:conditional}) and where language can optionally be used to facilitate learning (\S\ref{sec:assisted}).
In the former category we review instruction following, induction of reward from language, and environments with text in the action or observation space, all of which have language in the problem formulation itself.
In the latter, we review work that uses language to facilitate RL by transfer from domain-specific textual resources, or as a means of representing policies.

We conclude by identifying what we believe are the most important challenges for integrating natural language in RL (\S\ref{sec:conclusion}). 
Inspired by gaps in the existing literature, we advocate the development of new research environments utilizing domain knowledge in natural language, as well as a wider use of \gls{NLP} methods such as pre-trained language models and parsers to inform RL agents about the structure of the world.

\section{Background}
\label{sec:background}
\subsection{Reinforcement and Imitation Learning}

Reinforcement Learning~\citep{sutton2018reinforcement} is a framework that enables agents to reason about sequential decision making as an optimization process. Problems are formulated as Markov Decision Processes (MDPs), tuples $\langle S, A, T, R, \gamma \rangle$ where $S$ is the set of states, $A$ the set of actions, $T$ the transition probability function $T: S \times A \times S \rightarrow [0,1]$, $R$ the reward function $R: S \times A \times S \rightarrow \mathbb{R}$, and $\gamma \in [0,1)$ is a discount factor, typically set by either the environment or the agent designer. Given this setup, the goal of the optimization process is to find a policy $\pi(a | s) = p(A=a | S=s)$ that maximizes the expected discounted cumulative return $\sum_{k=0}^{\infty}{\gamma^{k} r_{k + 1}}$.
This framework is also used in Imitation Learning (IL), a setting in which the rewards are not observed, but the learning algorithm has access to a set of trajectories under optimal or sub-optimal policies. IL methods can then find approximations of the optimal policy, which can be used %
as a form of initialization, auxiliary objective, or for value estimation.

Since their inception, \gls{RL} algorithms have been successful in applications such as continuous control~\citep{white1992handbook}, dialogue systems~\citep{singh2002optimizing}, and board games~\citep{tesauro1995temporal}. Recent improvements in function approximation and pattern recognition made possible by deep learning have allowed RL to scale to problems with high dimensional input spaces such as video games~\citep{torrado2018deep} and complex planning problems such as Go~\citep{silver2017mastering}. Nonetheless, these methods remain sample inefficient, requiring millions or billions of interactions, and often generalize poorly to tasks only slightly different from those seen during training.
This severely limits the use of \gls{RL} for real-world tasks. See \citet{sutton2018reinforcement} for a comprehensive introduction to \gls{RL} and to \citet{arulkumaran_brief_2017} and \citet{osa2018algorithmic}  for reviews on recent algorithmic developments.

\subsection{Transfer from Natural Language}
\gls{NLP} has seen a recent surge of models that transfer syntactic and semantic knowledge to various downstream tasks. 
Current \gls{NLP} systems commonly employ deep learning models and embed (sequences of) words using dense vector representations. These vector representations are often pre-trained from large textual corpora and fine-tuned for a given task. Common techniques learn individual word representations from co-occurrence statistics
\citep{deerwester1990indexing,mikolov_distributed_2013} or contextual word-representations using (pseudo) language model objectives \citep{DBLP:conf/naacl/PetersNIGCLZ18,devlin_bert:_2018}. Both classes of models are motivated by Firth's distributional hypothesis (``You shall know a word by the company it keeps'') \citep{firth1957synopsis}, which suggests that the learned vector representation of a word like `scorpion' should be similar to `spider' if the corresponding words appear in similar contexts, \eg{}, if they can both be found around other words like `venomous' or `exoskeleton'.

Such (contextual) word representations can transfer knowledge to downstream tasks that have to deal with language as, for example, in \citep{socher_zero-shot_2013,frome_devise:_2013,DBLP:conf/acl/RuderH18,DBLP:conf/emnlp/PetersNZY18,DBLP:journals/corr/abs-1901-05287,tenney2018what}, to name just a few.
For instance, consider a text classification problem where we are tasked with assigning a document containing the word `scorpion' to the topic `arachnids' even though we have only observed `spider' during training.

World and task-specific knowledge communicated in natural language could be similarly transferred to sequential decision making problems.
For instance, learning agents can benefit from understanding explicit goals (``go to the door on the far side of the room''), constraints on policies (``avoid the scorpion''), or generic information about the reward or transition function (``scorpions are fast'').
Furthermore, pre-trained language models could play an important role in transferring world knowledge such as object affordances (``a key is used for [opening doors$\,|\,$unlocking boxes$\,|\,$investigating locked chests]'').
Similarly to recent question-answering \citep{DBLP:conf/acl/ChenFWB17} and dialog systems \citep{DBLP:journals/corr/abs-1811-01241}, agents could learn to make use of information retrieval and \gls{NLP} components to actively seek information required for making progress on a given task \citep{branavan_learning_2012}. %

\section{Current Use of Natural Language in RL}
\label{sec:nlp4rl}

In reviewing efforts that integrate language in RL we highlight work that develops tools, approaches, or insights that we believe may be particularly valuable for improving the generalization or sample efficiency of learning agents through the use of natural language.
As illustrated in Figure \ref{fig:illustration}, we separate the literature into \textbf{language-conditional} RL (in which interaction with language is necessitated by the problem formulation itself) and \textbf{language-assisted} RL (in which language is used to facilitate learning). The two categories are not mutually exclusive, in that for some language-conditional RL tasks, NLP methods or additional textual corpora are used to assist learning \citep{bahdanau_learning_2018,goyal_using_2019}.

To easily acquire data and constrain the difficulty of problems considered, the majority of these works use synthetic language (automatically generated from a simple grammar and limited vocabulary) rather than language generated by humans.
These often take the form of simple templates, \eg{} ``what colour is <object> in <room>'', but can be extended to more complex templates with relations and multiple clauses \citep{chevalier-boisvert_babyai:_2018}.

\subsection{Language-conditional RL}
\label{sec:conditional}

We first review literature for tasks in which integrating natural language is unavoidable, \ie{}, when the task itself is to interpret and execute instructions given in natural language, or natural language is part of the state and action space.
We argue in (\S\ref{sec:wild_corpora}) that approaches to such tasks can also be improved by developing methods that enable transfer from general and task-specific textual corpora. 
Methods developed for language-conditional tasks are relevant for language-assisted RL as they both deal with the problem of grounding natural language sentences in the context of RL.
Moreover, in tasks such as following sequences of instructions, the full instructions are often not necessary to solve the underlying RL problem but they assist learning by structuring the policy \citep{andreas_modular_2016} or by providing auxiliary rewards \citep{goyal_using_2019}.

\subsubsection{Instruction Following}
\label{sec:instruction}
Instruction following agents are presented with tasks defined by high-level (sequences of) instructions. We focus on instructions that are represented by (at least somewhat natural) language, and may take the form of formal specifications of appropriate actions, of goal states (or goals in general), or of desired policies. Effective instruction following agents execute the low level actions corresponding to the optimal policy or reach the goal specified by their instructions, and can generalize to unseen instructions during testing.

In a typical instruction following problem, the agent is given a description of the goal state or of a preferred policy as a proxy for a description of the task \citep{macmahon2006walk,kollar2010toward}.
Some work in this area focuses on simple object manipulation tasks~\citep{wang2016learning,bahdanau_learning_2018}, while other work focuses on 2D or 3D navigation tasks where the goal is to reach a specific entity.  
Entities might be described by predicates (``Go to the red hat")~\citep{hermann_grounded_2017,chaplot_gated-attention_2018}
or in relation to other entities (``Reach the cell above the westernmost rock.")~\citep{janner_representation_2017,chen2018touchdown}.  
Earlier approaches use object-level representation and relational modeling to exploit the structure of the instruction in relation to world entities, parsing the language instruction into a formal language ~\citep{kuhlmann2004guiding,chen2011learning,artzi_weakly_2013,andreas_alignment-based_2015}.
More recently, with the developments in deep learning, a common approach has been to embed both the instruction and observation to condition the policy directly~\citep{mei_listen_2015,hermann_grounded_2017,chaplot_gated-attention_2018,janner_representation_2017,misra_mapping_2017,chen2018touchdown}. 
Human-generated natural language instructions are used in \citet{macmahon2006walk,bisk2016natural,misra_mapping_2017,janner_representation_2017,chen2018touchdown,anderson2018vision,goyal_using_2019,wang_reinforced_2019}. Due to data-efficiency limitations of RL, this is not a standard in RL-based research \citep{hermann_grounded_2017}.

The line of work involving sequences of instructions has strong ties to Hierarchical RL \citep{barto2003recent}, with individual sentences or clauses from instructions corresponding to subtasks \citep{branavan2010reading}.
When the vocabulary of instructions is sufficiently simple, an explicit options policy can be constructed that associates each task description with its own modular sub-policy \citep{andreas_modular_2016}. A more flexible approach is to use a single policy that conditions on the currently executed instruction, allowing some generalization to unseen instructions 
\citep{mei_listen_2015,oh2017zero}. 
However, current approaches of this form require first pre-training the policy to interpret each of the primitives in a single-sentence instruction following setting. 
The internal compositional structure of instructions can then be exploited in various ways. 
For example, \citet{oh2017zero} achieve generalization to unseen instructions by forcing instruction embeddings to capture analogies, \eg{},~[Visit,X]~:~[Visit,Y]~::~[Pick up,X]~:~[Pick up,Y].

\subsubsection{Rewards from Instructions}
Another use of instructions is to induce a reward function for RL agents or planners to optimize.
This is relevant when the environment reward is not available to the agent at test time, but is either given during training~\citep{tellex2011understanding} or can be inferred from (parts of) expert trajectories. In order to apply instruction following more broadly, there needs to be a way to automatically evaluate whether the task specified by the instruction has been completed.
The work addressing this setting is influenced by methods from the inverse reinforcement learning (IRL) literature~\citep{ziebart_maximum_2008,ho_generative_2016}.
A common architecture consists of a reward-learning module that learns to ground an instruction to a (sub-)goal state or trajectory segment, and is used to generate a reward for a policy-learning module or planner.

When full demonstrations are available, the reward function can be learned using standard IRL methods like MaxEnt IRL \citep{ziebart_maximum_2008} as in \cite{fu_language_2018}, or maximum likelihood IRL \citep{babes2011apprenticeship} as in \cite{macglashan_grounding_2015}, who also learn a joint generative model of rewards, behaviour, and language.
Otherwise, given a dataset of goal-instruction pairs, as in \cite{bahdanau_learning_2018}, the reward function is learned through an adversarial process similar to that of \cite{ho_generative_2016}. For a given instruction, the reward-learning module aims to discriminate goal states from the states visited by the policy (assumed non-goal), while the agent is rewarded for visiting states the discriminator cannot distinguish from the goal states.

When environment rewards are available but sparse, instructions may still be used to generate auxiliary rewards to help learn efficiently.
In this setting, \cite{goyal_using_2019} and \cite{wang_reinforced_2019} use auxiliary reward-learning modules trained offline to predict whether trajectory segments correspond to natural language annotations of expert trajectories.
\cite{agarwal_learning_2019} perform a meta-optimisation to learn auxiliary rewards conditioned on features extracted from instructions. The auxiliary rewards are learned so as to increase performance on the true objective after being used for a policy update.
As some environment rewards are available, these settings are closer to language-assisted RL.

\subsubsection{Language in the Observation and Action Space}
\label{sec:games}
Environments that use natural language as a first-class citizen for driving the interaction with the agent present a strong challenge for RL algorithms.
Using natural language requires common sense, world knowledge, and context to resolve ambiguity and cheaply encode information \cite{mey_pragmatics:_1993}.
Furthermore, linguistic observation and action spaces grow combinatorially as the size of the vocabulary and the complexity of the grammar increase. 
For instance, compare the space of possible instructions when following cardinal directions (\eg{} ``go north'') with reaching a position that is described in relative terms (\eg{} ``go to the blue ball south west of the green box'').

Text games, such as Zork \cite{zork1980}, are easily framed as RL environments and make a good testbed for structure learning, knowledge extraction, and transfer across tasks~\citep{branavan_learning_2012}. %
\citep{depristo2001being,narasimhan2015language,yuan_counting_2018} observe that when the action space of the text game is constrained to verb-object pairs, decomposing the $Q$-function into separate parts for verb and object provides enough structure to make learning more tractable.
However, they do not show how to scale this approach to action-sentences of arbitrary length. 
To facilitate the development of a consistent set of benchmarks in this problem space, \cite{cote_textworld:_2018} propose \emph{TextWorld}, a framework that allows the generation of instances of text games that behave as RL environments.
They note that existing work on word-level embedding models for text games (e.g. ~\citep{kostka_text-based_2017,fulda2017can})
 achieve good performance only on easy tasks.
 
Other examples of settings where agents are required to interact using language include \emph{dialogue systems} and \emph{question answering} (Q\&A).
The two have been a historical focus in \gls{NLP} research and are extensively reviewed by \cite{chen2017survey} and \cite{bouziane2015question} respectively.
Recent work on visual Q\&A (VQA) have produced a first exploration of multi-modal settings in which agents are tasked with performing both visual and language-based reasoning~\citep{antol2015vqa,johnson2017clevr,massiceti2018flipdial}. Embodied Q\&A (EQA) extends this setting, by requiring agent to explore and navigate the environment in order to answer queries ~\citep{das2018embodied,gordon2018iqa}, for example, ``How many mugs are in the kitchen?'' or ``Is there a tomato in the fridge?''. By employing rich 3D environments, EQA tasks require agents to carry out multi-step planning and reasoning under partial observability.
However, since in the existing work the agents only choose from a limited set of short answers instead of generating an arbitrary length response, so far such tasks have been very close to the instruction following setting.

\subsection{Language-assisted RL}
\label{sec:assisted}

In this section, we consider work that explores how knowledge about the structure of the world can be transferred from natural language corpora and methods into RL tasks, in cases where language itself is not essential to the task. 
Textual information can assist learning by specifying informative features,
annotating states or entities in the environment,
or describing subtasks in a multitask setting. 
In most cases covered here, the textual information is task-specific, with a few cases of using task-independent information through language parsers \citep{branavan_learning_2012} and pre-trained sentence embeddings \citep{goyal_using_2019}.

\subsubsection{Language for Communicating Domain Knowledge} 
\label{sec:domain}
In a more general setting than instruction following, any kind of text containing potentially task-relevant information could be available. 
Such text may contain advice regarding the policy an agent should follow or information about the environment dynamics (see \Cref{fig:illustration}).
Unstructured and descriptive (in contrast to instructive) textual information is more abundant and can be found in wikis, manuals, books, or the web.
However, using such information requires (i) retrieving useful information for a given context and (ii) grounding that information with respect to observations. 

\citet{eisenstein_reading_2009} learn abstractions in the form of conjunctions of predicate-argument structures that can reconstruct sentences and syntax in task-relevant documents using a generative language model. These abstractions are used to obtain a feature space that improves imitation learning outcomes. %
\cite{branavan_learning_2012} learn a $Q$-function that improves Monte-Carlo tree search planning for the first few moves in Civilization II, a turn-based strategy game, while accessing the game's natural language manual.
Features for their $Q$-function depend on the game manual via learned sentence-relevance and predicate-labeling models whose parameters are optimised only by minimising the $Q$-function estimation error.
Due to the structure of their hand-crafted features (some of which match states and actions to words, \eg{} \texttt{action == irrigate AND action-word == "irrigate"}) , these language processing models nonetheless somewhat learn to extract relevant sentences and classify their words as relating to the state, action, or neither.
More recently, \cite{narasimhan_grounding_2017} investigate planning in a 2D game environment where properties of entities in the environment are annotated by natural language (\eg{} the `spider' and `scorpion' entities might be annotated with the descriptions ``randomly moving enemy" and ``an enemy who chases you", respectively).
Descriptive annotations facilitate transfer by learning a mapping between the annotations and the transition dynamics of the environment.

\subsubsection{Language for Structuring Policies}  
\label{sec:representation} 

One use of natural language is communicating information about the state and/or dynamics of an environment.
As such it is an interesting candidate for constructing priors on the model structure or representations of an agent. 
This could include shaping representations towards more generalizable abstractions, making the representation space more interpretable to humans, or efficiently structuring the computations within a model.

\cite{andreas2016neural} propose a neural architecture that is dynamically composed of a collection of jointly-trained neural modules, based on the parse tree of a natural language prompt. 
While originally developed for visual question answering, \cite{das_neural_2018} and \cite{bahdanau_learning_2018} successfully apply variants of this idea to RL tasks.  
\cite{andreas_learning_2017} explore the idea of natural language descriptions as a policy parametrization in a 2D navigation task adapted from \cite{janner_representation_2017}. In a pre-training phase, the agent learns to imitate expert trajectories conditioned on instructions. By searching in the space of natural language instructions, the agent is then adapted to the new task where instructions and expert trajectories are not available.  
  
The hierarchical structure of natural language and its compositionality make it a particularly good candidate for representing policies in hierarchical \gls{RL}. %
\cite{andreas_modular_2016} and \cite{shu_hierarchical_2017} can be viewed as using language (rather than logical or learned representations) as policy specifications for a hierarchical agent. 
More recently, \cite{hu2019hierarchical} consider generated natural language as a representation for macro actions in a real-time strategy game environment based on \cite{tian_elf:_2017}. In an IL setting, a meta-controller is trained to generate a sequence of natural language instructions. Simultaneously, a base-controller policy is trained to execute these generated instructions through a sequence of actions.

\section{Trends for Natural Language in RL}
\label{sec:conclusion}

The preceding sections surveyed the literature exploring how natural language can be integrated with \gls{RL}. 
Several trends are evident:
(i) studies for language-conditional RL are more numerous than for language-assisted RL, 
(ii) learning from task-dependent text is more common than learning from task-independent text,
(iii) within work studying transfer from task-dependent text, only a handful of papers study how to use unstructured and descriptive text,
(iv) there are only a few papers exploring methods for structuring internal plans and building compositional representations using the structure of language,
and finally (v) natural language, as opposed to synthetically generated languages, is still not the standard in research on instruction following. 

To advance the field, we argue that more research effort should be spent on learning from naturally occurring text corpora in contrast to instruction following. 
While learning from unstructured and descriptive text is particularly difficult, it has a much greater application range and potential for impact. %
Moreover, we argue for the development of more diverse environments with real-world semantics. 
The tasks used so far use small and synthetic language corpora and are too artificial to significantly benefit from transfer from real-world textual corpora. 
In addition, we emphasize the importance of developing standardized environments and evaluations for comparing and measuring progress of models that integrate natural language into \gls{RL} agents.

We believe that there are several factors that make focusing such efforts worthwhile now: (i) recent progress in pre-training language models, (ii) general advances in representation learning, as well as (iii) development of tools that make constructing rich and challenging RL environments easier. 
Some significant work, especially in language-assisted RL, has been done prior to the surge of deep learning methods \citep{eisenstein_reading_2009,branavan_learning_2012}, and is worth revisiting. 
In addition, we encourage the reuse of software infrastructure, e.g. \citep{bahdanau_learning_2018,cote_textworld:_2018,chevalier-boisvert_babyai:_2018} for constructing environments and standardized tests.

\subsection{Learning from Text Corpora in the Wild}
\label{sec:wild_corpora}

The web contains abundant textual resources that provide instructions and how-to's.\footnote{\eg{} \url{https://www.wikihow.com/} or \url{https://stackexchange.com/}}
For many games (which are often used as testbeds for RL), detailed walkthroughs and strategy guides exist.
We believe that transfer from task-independent corpora could also enable agents to better utilize such task-dependent corpora. 
Preliminary results that demonstrate zero-shot capabilities \citep{radford2019language} suggest that a relatively small dataset of instructions or descriptions could suffice to ground and consequently utilize task-dependent information for better sample efficiency and generalization of \gls{RL} agents.

\subsubsection{Task-independent Corpora}

Natural language reflects human knowledge about the world \citep{Zellers2018SWAG}.
For instance, an effective language model should assign a higher probability to ``get the green apple from the tree behind the house" than to ``get the green tree from the apple behind the house". 
Harnessing such implicit commonsense knowledge captured by statistical language models could enable transfer of knowledge to \gls{RL} agents. 

In the short-term, we anticipate more use of pre-trained word and sentence representations for research on language-conditional RL. 
For example, consider instruction following with natural language annotations. 
Without transfer from a language model (or another language grounding task as in \cite{yu_interactive_2018}), instruction-following systems cannot generalize to instructions outside of the training distribution containing unseen synonyms or paraphrases (\eg{} ``fetch a stick", ``return with a stick", ``grab a stick and come back"). 
While pre-trained word and sentence representations alone will not solve the problem of grounding an unseen object or action, they do help with generalization to instructions with similar meaning but unseen words and phrases. %
In addition, we believe that learning representations for transferring knowledge about analogies, going beyond using analogies as auxiliary tasks \citep{oh2017zero} will play an important role in generalizing to unseen instructions. %

As pre-trained language models and automated question answering systems become more capable, %
one interesting long-term direction %
is studying agents that can query knowledge more explicitly. 
For example, during the process of planning in natural language, an agent that has a pre-trained language model as sub-component could let the latter complete ``to open the door, I need to..." with ``turn the handle". Such an approach could be expected to learn more rapidly than \emph{tabula rasa} reinforcement learning. 
However, such agents would need to be capable of reasoning and planning in natural language, which is a related line of work (see Language for Structuring Policies in \S\ref{sec:representation}).

\subsubsection{Task-dependent Corpora} 

Research on transfer from descriptive task-dependent corpora is promising due to its wide application potential. 
It also requires development of new environments, as early research may require access to relatively structured and partially grounded forms of descriptive language similarly to \cite{narasimhan_grounding_2017}. 
One avenue for early research is developing environments with relatively complex but still synthetic languages, providing information about environmental dynamics or advice about good strategies.
For example, in works studying transfer from descriptive task-dependent language corpora \citep{narasimhan_grounding_2017}, natural language sentences could be embedded using representations from pre-trained language models. 
Integrating and fine-tuning pre-trained information retrieval and machine reading systems similar to \citep{DBLP:conf/acl/ChenFWB17} with RL agents that query them could help in extracting and utilizing relevant information from unstructured task-specific language corpora such as game manuals as used in \citet{branavan_learning_2012}.

\subsection{Towards Diverse Environments with Real-World Semantics}
\label{sec:real_environments}
One of the central promises of language in \gls{RL} is the ability to rapidly specify and help agents adapt to new goals, reward functions, and environment dynamics. This capability is not exercised at all by standard RL benchmarks like strategy games (which typically evaluate agents against a single or small number of fixed reward functions). It is evaluated in only a limited way by existing instruction following benchmarks, which operate in closed-task domains (navigation, object manipulation, etc.) and closed worlds. The simplicity of these tasks is often reflected in the simplicity of the language that describes them, with small vocabulary sizes and multiple pieces of independent evidence for the grounding of each word.

Real natural language has important statistical properties, such as the power-law distribution of word frequencies \citep{zipf1949human} which does not appear in environments with synthetically generated language and small numbers of entities. Without environments that encourage humans to process (and force agents to learn from) complex composition and the ``long tail'' of lexicon entries, we cannot expect to hope that \gls{RL} agents will generalize at all outside of \emph{closed-world tasks}.

One starting point is provided by extending 3D house simulation environments \citep{gordon2018iqa,das2018embodied,yan2018chalet}
some of which already support generation and evaluation of templated instructions. These environments contain more real-world semantics (e.g. a microwave can be opened, a car is in the garage).
However, interactions with the objects and the available templates for language instructions have been very limited so far.
Another option is presented by open-world video games like Minecraft \citep{johnson2016malmo}, in which users are free to assemble complex structures from simple parts, and thus have an essentially unlimited universe of possible objects to describe and goals involving those objects. 
More work is needed in exploring how learning grounding scales with the number of human annotations and environmental interactions \citet{bahdanau_learning_2018,chevalier-boisvert_babyai:_2018}.
Looking ahead, as core machine learning tools for learning from feedback and demonstrations become sample-efficient enough to use in the real world, we anticipate that approaches combining language and \gls{RL} will find applications as wide-ranging as autonomous vehicles, virtual assistants and household robots.

\section{Conclusion}

The currently predominant way \gls{RL} agents are trained restricts their use to environments where all information about the policy can be gathered from directly acting in and receiving reward from the environment. 
This \emph{tabula rasa} learning results in low sample efficiency and poor performance when transferring to other environments.
Utilizing natural language in \gls{RL} agents could drastically change this by transferring knowledge from natural language corpora to \gls{RL} tasks, as well as between tasks, consequently unlocking \gls{RL} for more diverse and real-world tasks.
While there is a growing body of papers that incorporate language into \gls{RL}, %
most of the research effort has been focused on simple \gls{RL} tasks and synthetic languages, with highly structured and instructive text. 

To realize the potential of language in \gls{RL}, we advocate for more research into learning from unstructured or descriptive language corpora, with a greater use of \gls{NLP} tools like pre-trained language models. 
Such research also requires development of more challenging environments that reflect the semantics and diversity of the real world.

\section*{Acknowledgements}
This project has received funding from the European Research Council (ERC) under the European Union’s Horizon 2020 research and innovation programme (grant agreement number 637713). JL has been funded by EPSRC Doctoral Training Partnership and Oxford-DeepMind Scholarship, and GF has been funded by UK EPSRC CDT in Autonomous Intelligent Machines and Systems. We also thank Miles Brundage, Ethan Perez and Sam Devlin for feedback on the paper draft.

\bibliographystyle{named}
\bibliography{bib/nlp4rl,bib/refs_tim}

\end{document}